\documentclass{article} 
\usepackage{iclr2021_conference,times}

\usepackage{amsmath,amsfonts,bm}

\def\eqref#1{equation~\ref{#1}}

\def\1{\bm{1}}

\DeclareMathAlphabet{\mathsfit}{\encodingdefault}{\sfdefault}{m}{sl}
\SetMathAlphabet{\mathsfit}{bold}{\encodingdefault}{\sfdefault}{bx}{n}

\usepackage{hyperref}
\usepackage{url}
\usepackage{amsthm}
\usepackage{amssymb}
\usepackage{graphicx}
\usepackage{comment}
\usepackage{xcolor}
\usepackage{listings}

\lstdefinestyle{mystyle}{
	keywordstyle=\bfseries,
	keywords={ge,return,def,not,xor,ne},
}
\title{$\partial\mathbb{B}$ nets: learning discrete functions by\\gradient descent}

\author{Ian Wright\thanks{wrighti@acm.org}}

\newtheorem{theorem}{Theorem}
\newtheorem*{definition}{Definition}
\newtheorem{prop}{Proposition}
\newtheorem{lemma}{Lemma}

\makeatletter

\DeclareRobustCommand{\btleft}{\mathrel{\mathpalette\btlr@\blacktriangleleft}}
\DeclareRobustCommand{\btright}{\mathrel{\mathpalette\btlr@\blacktriangleright}}

\newcommand{\btlr@}[2]{%
	\begingroup
	\sbox\z@{$\m@th#1\triangleright$}%
	\sbox\tw@{\resizebox{1.1\wd\z@}{1.1\ht\z@}{\raisebox{\depth}{$\m@th#1\mkern-1mu#2$}}}%
	\ht\tw@=\ht\z@ \dp\tw@=\dp\z@ \wd\tw@=\wd\z@
	\copy\tw@
	\endgroup
}

\iclrfinalcopy 
\begin{document}

\maketitle

\begin{abstract}
	$\partial\mathbb{B}$ nets are differentiable neural networks that learn discrete boolean-valued functions by gradient descent. $\partial\mathbb{B}$ nets have two semantically equivalent aspects: a differentiable soft-net, with real weights, and a non-differentiable hard-net, with boolean weights. We train the soft-net by backpropagation and then `harden' the learned weights to yield boolean weights that bind with the hard-net. The result is a learned discrete function. `Hardening' involves no loss of accuracy, unlike existing approaches to neural network binarization. Preliminary experiments demonstrate that $\partial\mathbb{B}$ nets achieve comparable performance on standard machine learning problems yet are compact (due to 1-bit weights) and interpretable (due to the logical nature of the learnt functions).
\end{abstract}

\section{Introduction}

Neural networks are differentiable functions with weights represented by machine floats. Networks are trained by gradient descent in weight-space, where the direction of descent minimises loss. The gradients are efficiently calculated by the backpropagation algorithm \citep{rumelhart1986learning}. This overall approach has led to tremendous advances in machine learning.

However, there are drawbacks. First, differentiability means we cannot directly learn discrete functions, such as logical predicates. In consequence, what a network has learned is difficult to interpret and verify. Second, representing weights as machine floats enables time-efficient training but at the cost of memory-inefficient models. For example, network quantisation techniques (see \cite{QIN2020107281}) demonstrate that full 64 or 32-bit precision weights are often unnecessary for final predictive performance, although there is a trade-off.

\begin{figure}[h]
	\centering
	\includegraphics[width=0.85\textwidth]{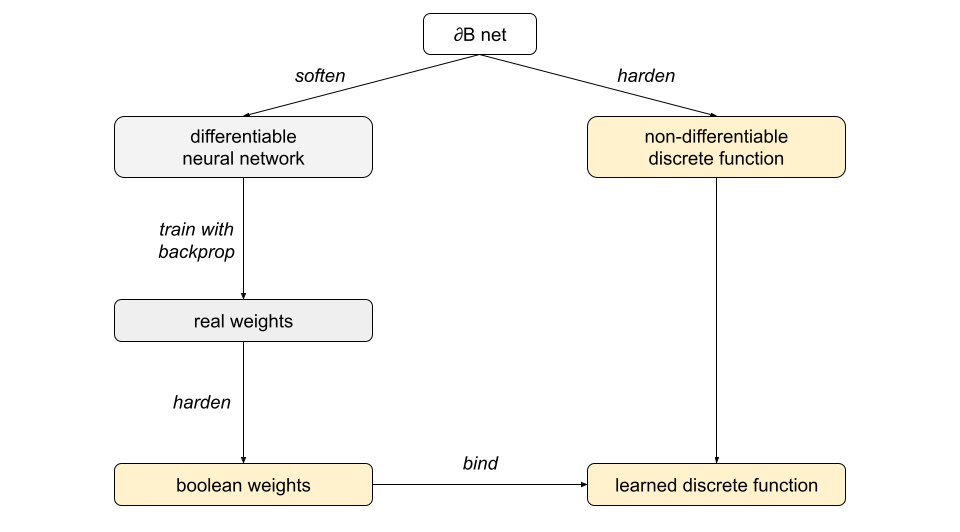}
	\caption{{\em Learning discrete functions with a $\partial\mathbb{B}$ net.} A $\partial \mathbb{B}$ net specifies (i) a differentiable neural network that is hard-equivalent to (ii) a non-differentiable discrete function. The neural network is trained as normal with backpropagation to yield a set of real weights. The real weights are hardened to boolean values and then bound with the discrete function. The result is a learned discrete function that performs identically to the trained network.}
	\label{fig:main-idea}
\end{figure}

A standard approach to mitigate these drawbacks is to approximate discrete functions by defining continuous relaxations. This paper explores a different approach: we define differentiable functions that `harden', without approximation, to discrete functions. Specifically, we define $\partial \mathbb{B}$ nets that have two equivalent aspects: a {\em soft-net}, which is a differentiable real-valued function, and a {\em hard-net}, which is a non-differentiable, discrete function. Both aspects are semantically equivalent. We train the soft-net as normal, using backpropagation, then `harden' the learned weights to boolean values, which we then bind with the hard-net to yield a discrete function with identical predictive performance (see figure \ref{fig:main-idea}). In consequence, interpreting and verifying a $\partial \mathbb{B}$ net is relatively less difficult. And boolean-valued, 1-bit weights significantly increase the memory-efficiency of trained models.

The main contributions of this work are (i) defining novel activation functions that `harden' to semantically equivalent discrete functions, (ii) defining novel network architectures to effectively learn discrete functions that solve multi-class classification problems, and (iii) experiments that demonstrate $\partial \mathbb{B}$ nets compete with existing approaches in terms of predictive performance yet yield compact models.

Section \ref{sec:related-work} discusses related work, section \ref{sec:db-nets} defines $\partial\mathbb{B}$ nets, section \ref{sec:experiments} presents experimental results, and section \ref{sec:conclusion} concludes.

\section{Related work}\label{sec:related-work}

Methods that learn discrete boolean functions can be broadly categorized as either non-differentiable or differentiable.

Non-differentiable approaches include boolean-valued decision trees \citep{BreiFrieStonOlsh84}, random forests \citep{598994}, genetic programming \citep{koza1992genetic} and, more recently, Tsetlin machines \cite{granmo18}. Tsetlin machines represent propositional formulae by collections of simple automata with integer weights optimised by positive and negative feedback defined in terms of a hard threshold function. These models directly represent boolean decisions and therefore are easier to interpret compared to deep neural networks. However, they tend to perform less well, compared to differentiable approaches such as deep learning, on large volumes of high-dimensional data (e.g. NLP, images and audio) without manual feature engineering, although Tsetlin machines show promise on such tasks \citep{Granmo2019TheCT}.

Differentiable approaches that learn boolean functions include (i) systems that integrate rule-based reasoning with neural components, and (ii) binarization techniques that quantize neural networks by converting real-valued weights and activations to binary values. For example, differentiable inductive logic programming \citep{10.5555/3241691.3241692}, neural logic machines \citep{dong2018neural} 
and differentiable neural logic networks \cite{DBLP:phd/basesearch/Payani20} learn first-order logic rules using gradient descent. These systems combine the benefits of logical inference and interpretability with end-to-end differentiability. However, they rely on combinatoric enumeration of rulesets and therefore do not scale to large datasets. The technique of network binarization aims to significantly reduce model size and inference costs while maintaining predictive accuracy. Binarization reduces a real-valued neural network to a binary network where nonlinear activation functions are replaced by boolean majority functions. For example, BinaryConnect \citep{10.5555/2969442.2969588}, XNOR-Net \citep{10.1007/978-3-319-46493-0_32}, and LUTNet \citep{9026948}, optimize a continuous relaxation or approximation of the binary net during training. However, binary-valued functions are intrinsically non-differentiable and therefore training by gradient descent is challenging. Plus, binarization throws away information, which reduces accuracy \citep{QIN2020107281}. 

The design-space of algorithms that learn boolean functions is large, with various trade-offs. In this paper we investigate an under-explored area of differentiable nets that are semantically equivalent, without approximation or loss, to an arbitrarily complex boolean function. We aim to combine the benefits of deep neural networks trained by gradient descent with the efficiency, interpretability and logical bias of boolean functions -- but without loss of accuracy.

\section{$\partial\mathbb{B}$ nets}\label{sec:db-nets}

A $\partial \mathbb{B}$ net has two aspects, a soft-net and a hard-net. Both nets use bits to represent transitory values and learnable weights, but a soft-net uses soft-bits and a hard-net uses hard-bits.

\begin{definition}[Soft-bits and hard-bits]
A {\em soft-bit} is a real value in the range $[0,1]$ and a {\em hard-bit} is a boolean value from the set $\{0,1\}$. A soft-bit, $x$, is {\em high} if $x>1/2$, otherwise it is {\em low}.
\end{definition}

A hardening function converts soft-bits to hard-bits.

\begin{definition}[Hardening]
The {\em hardening} function, $\operatorname{harden}(x_{1}, \dots, x_{n}) = [f(x_{1}), \dots, f(x_{n})]$, converts soft-bits to hard-bits, where
\begin{equation*}
f(x) =
\begin{cases}
1 & \text{if } x > 1/2 \\
0 & \text{otherwise.}
\end{cases}
\end{equation*}
\end{definition}

The soft-bit value $1/2$ is therefore a threshold. Above this threshold the soft-bit represents $\text{True}$, otherwise it represents $\text{False}$.

A soft-net is any differentiable function, $f$, that `hardens' to a semantically equivalent discrete function, $g$. For example, if $f(x) = 1 - x$, where $x \in [0,1]$, and $g(y) = \neg y$, where $y \in \{0,1\}$ then: if $x$ is high (resp. low) then both $f(x)$ and $g(\operatorname{harden}(x))$ are low (resp. high). In other words, $f$ is hard-equivalent to boolean negation. More generally:

\begin{definition}[Hard-equivalence]
	A function, $f: [0,1]^n \rightarrow [0,1]^m$, is {\em hard-equivalent} to a discrete function, $g: \{1,0\}^n \rightarrow \{1,0\}^m$,	if
	\begin{equation*}
	\operatorname{harden}(f({\bf x})) = g(\operatorname{harden}({\bf x}))
	\end{equation*}
	for all ${\bf x} \in \{(x_{1}, \dots, x_{n}) ~|~ x_{i} \in [0,1] \setminus \{1/2\}\}$. For shorthand write $f \btright g$.
\end{definition}

Neural networks are typically composed of nonlinear activation functions (for representational generality) that are strictly monotonic (so gradients always exist that link changes in inputs to outputs without local minima) and smooth (so gradients reliably represent the local loss surface). However, activation functions that are monotonic but not strictly (so some gradients are zero) and differentiable almost everywhere (so some gradients are undefined) can also work, e.g. RELU \citep{10.5555/3104322.3104425}. $\partial \mathbb{B}$ nets are composed from `activation' functions that also satisfy these properties plus the additional property of hard-equivalence to a boolean function (and natural generalisations). We now turn to specifying the kind of `activation' functions used by $\partial \mathbb{B}$ nets.

\begin{figure}[t!]
	\centering
	\includegraphics[trim=0pt 0pt 0pt 0pt, clip, width=1.0\textwidth]{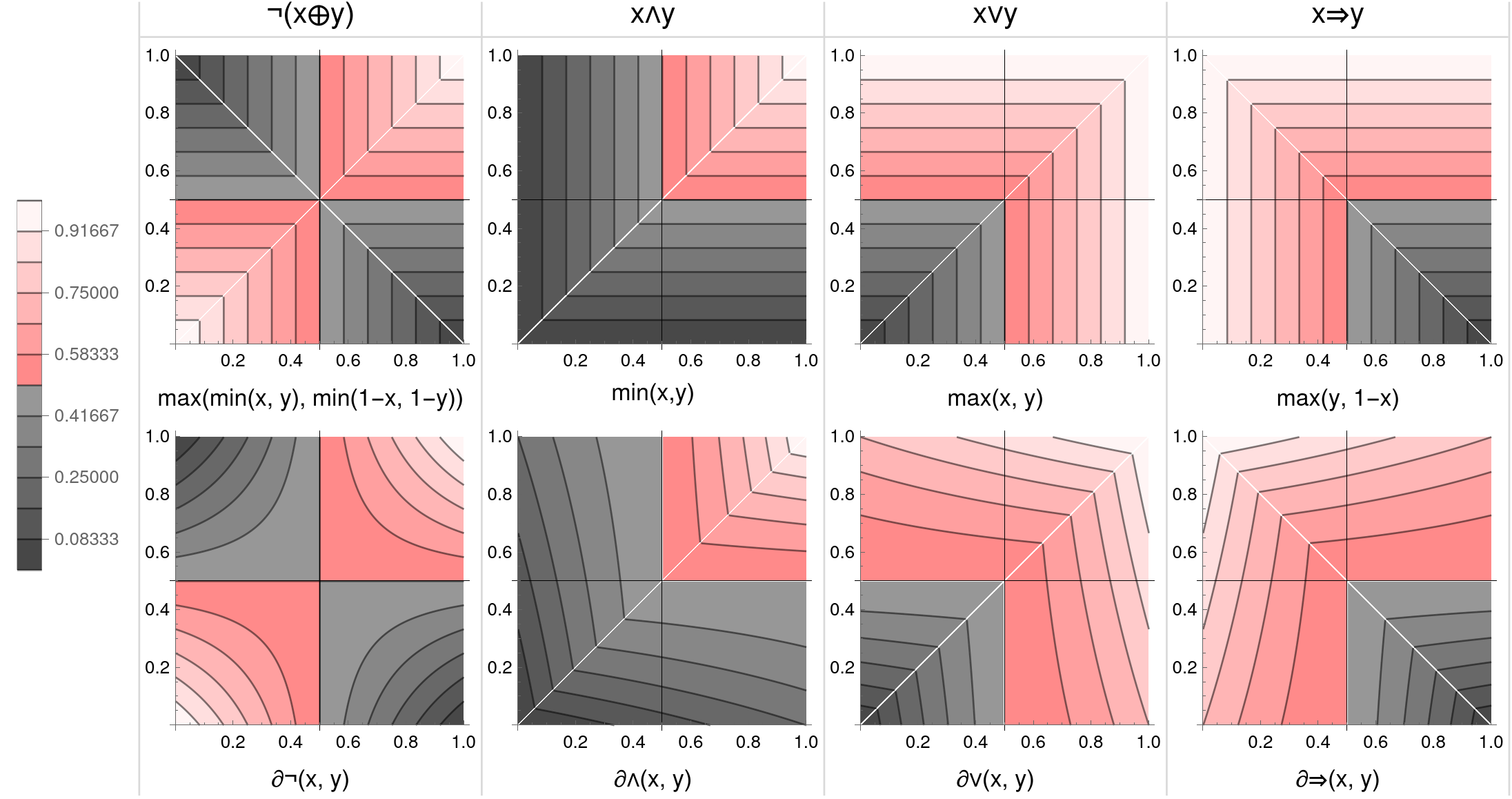}
	\caption{{\em Gradient-rich versus gradient-sparse differentiable boolean functions.} Each column contains contour plots of functions $f(x,y)$ that are hard-equivalent to a boolean function (one of $\neg(x \oplus y)$, $x \wedge y$, $x \vee y$, or $x \Rightarrow y$). Every function is continuous and differentiable almost everywhere (white lines indicate non-continuous derivatives). The upper plots are gradient-sparse, where vertical and horizontal contours indicate the function is constant with respect to one of its inputs, i.e. $\partial f/\partial y = 0$ or $\partial f/\partial x = 0$. The lower plots are gradient-rich, where the curved contours indicate the function always varies with respect to any of its inputs, i.e. $\partial f/\partial y \neq 0$ and $\partial f/\partial x \neq 0$. $\partial \mathbb{B}$ nets use gradient-rich functions to ensure that error is always backpropagated to all inputs.} 
	\label{fig:gradient-rich}
\end{figure}

\subsection{Learning to negate}

Say we aim to learn to negate a boolean value, $x$, or leave it unaltered. Represent this decision by a boolean weight, $w$, where low $w$ means negate and high $w$ means do not negate. The boolean function that meets this requirement is $\neg(x \oplus w)$. However, this function is not differentiable. Define the differentiable function,
	\begin{equation*}
	\begin{aligned}
	\partial_{\neg}: [0, 1]^{2} &\to [0,1], \\
	(w, x) &\mapsto 1 - w + x (2w - 1)\text{,}
	\end{aligned}
	\end{equation*}
where $\partial_{\neg}(w, x) \btright \neg(x \oplus w)$ (see proposition \ref{prop:not}).

There are many kinds of differentiable fuzzy logic operators (see \cite{VANKRIEKEN2022103602} for a review). So why this functional form? Product logics, where $f(x,y) = x y$ is as a soft version of $x \wedge y$, although hard-equivalent at extreme values, e.g. $f(1,1)=1$ and $f(0,1)=0$, are not hard-equivalent at intermediate values, e.g. $f(0.6, 0.6) = 0.36$, which hardens to $\operatorname{False}$ not $\operatorname{True}$. G\"{o}del-style $\operatorname{min}$ and $\operatorname{max}$ functions, although hard-equivalent over the entire soft-bit range, i.e. $\operatorname{min}(x,y) \btright x \wedge y$ and $\operatorname{max}(x,y) \btright x \vee y$, are gradient-sparse in the sense that their outputs do not always vary when any input changes, e.g. $\frac{\partial}{\partial x} \operatorname{max}(x,y) = 0$ when $(x,y)=(0.1, 0.9)$. So although the composite function $\operatorname{max}(\operatorname{min}(w, x), \operatorname{min}(1-w, 1-x))$ is differentiable and $\btright \neg(x \oplus w)$ it does not always backpropagate error to its inputs. In contrast, $\partial_{\neg}$ always backpropagates error to its inputs because it is a gradient-rich function (see figure \ref{fig:gradient-rich}). 

\begin{definition}[Gradient-rich]
	A function, $f: [0,1]^n \rightarrow [0,1]^m$, is {\em gradient-rich} if $\frac{\partial f({\bf x})}{\partial x_{i}} \neq {\bf 0}$ for all ${\bf x} \in \{(x_{1}, \dots, x_{n}) ~|~ x_{i} \in [0,1] \setminus \{1/2\}\}$.
\end{definition}

$\partial \mathbb{B}$ nets must be composed of `activation' functions that are hard-equivalent to discrete functions but also, where possible, gradient-rich. To meet this requirement we introduce the technique of margin packing.

\subsection{Margin packing}

Say we aim to construct a differentiable analogue of $x \wedge y$. Note that $\operatorname{min}(x,y)$ essentially selects one of $x$ or $y$ as a representative soft-bit that is guaranteed hard-equivalent to $x \wedge y$. However, by selecting only one of $x$ or $y$ then $\operatorname{min}$ is also guaranteed to be gradient-sparse. We define a `margin packing' method to solve this dilemma.

The main idea of margin packing is (i) select a representative bit that is hard-equivalent to the target discrete function, and then (ii) pack a fraction of the margin between the representative bit and the hard threshold $1/2$ with gradient-rich information. The result is an augmented bit that is a function of all inputs yet hard-equivalent to the target function.

\begin{figure}[t!]
	\centering
	\includegraphics[trim=30pt 5pt 30pt 10pt, clip, width=1.0\textwidth]{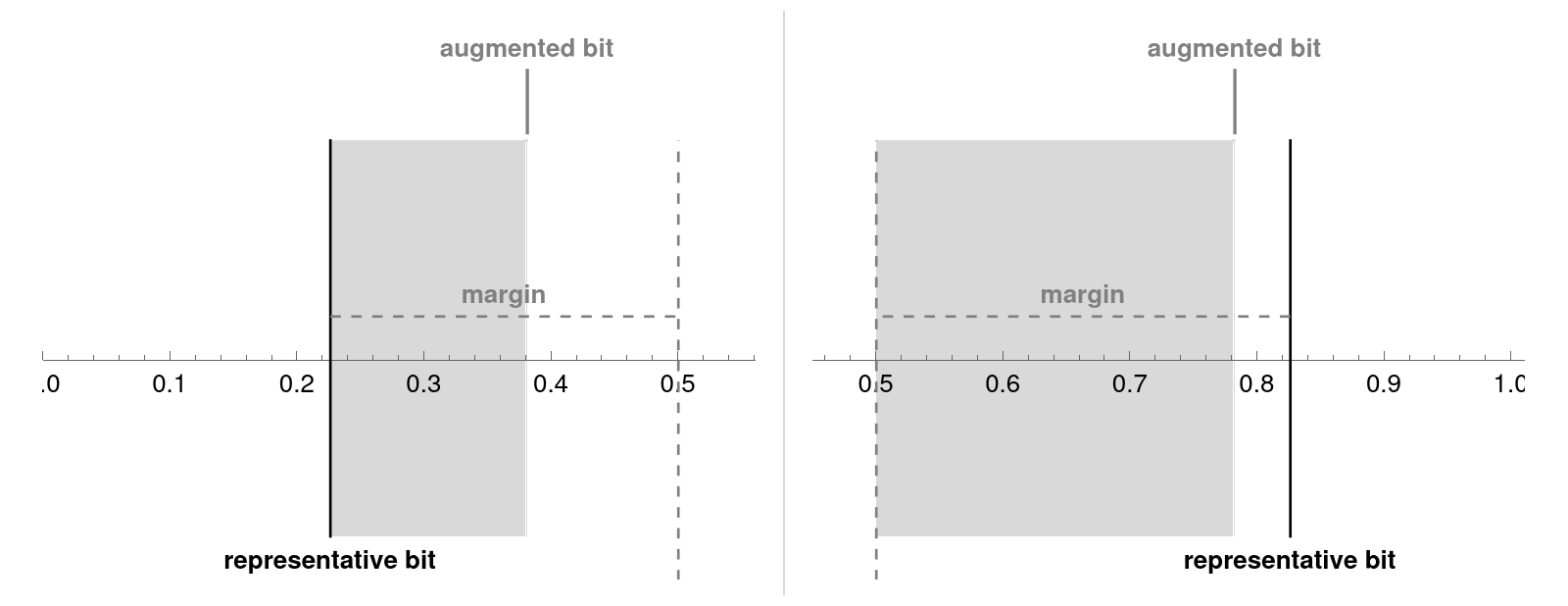}
	\caption{{\em Margin packing for constructing gradient-rich, hard-equivalent functions}. A representative bit, $z$, is hard-equivalent to a discrete target function but gradient-sparse (e.g. $z=\operatorname{min}(x,y) \btright x \wedge y$). On the left $z$ is low, $z<1/2$; on the right $z$ is high, $z>1/2$. We can pack a fraction of the margin between $z$ and the hard threshold $1/2$ with additional gradient-rich information without affecting hard-equivalence. A natural choice is the mean soft-bit, $\bar{\bf x} \in [0,1]$. The grey shaded areas denote the packed margins and the final augmented bit. On the left $\approx 60\%$ of the margin is packed; on the right $\approx 90\%$.}
	\label{fig:margin-trick}
\end{figure}

More concretely, say we have a vector of soft-bit inputs ${\bf x}$ and the $i$th element represents the target discrete function (e.g. if our target is $x \wedge y$ then ${\bf x}=[x,y]$ and $i$ is 1 if $x<y$ and $i=2$ otherwise). Now, if we pack only a fraction of the available margin, $|x_{i}-1/2|$, we will not cross the $1/2$ threshold and break the hard-equivalence of the representative bit. The average soft-bit value, $\bar{\bf x} \in [0,1]$, is just such a gradient-rich fraction. We therefore define 
\begin{equation*}
\begin{aligned}
\operatorname{margin-fraction}: [0,1]^{n} \times 1,2,\dots,n &\to [0,1],\\
({\bf x}, i) &\mapsto \bar{\bf x} \times \left|x_{i} - 1/2\right| \text{.}
\end{aligned}
\end{equation*}
The packed fraction, $\bar{\bf x}$, of the margin increases or decreases with the average soft-bit value. The available margin, $\left|x_{i} - 1/2\right|$, tends to zero as the representative bit, $x_{i}$, tends to the hard threshold $1/2$. At the threshold point there is no margin to pack. Now, define the augmented bit as
\begin{equation}
\begin{aligned}
\operatorname{augmented-bit}: [0,1]^{n} \times 1,2,\dots,n &\to [0,1],\\
({\bf x}, i) &\mapsto 
\begin{cases}
1/2 + \operatorname{margin-fraction}({\bf x}, i) & \text{if } x_{i} > 1/2 \\
x_{i} + \operatorname{margin-fraction}({\bf x}, i) & \text{otherwise.}
\end{cases}
\end{aligned}
\label{eq:augmented-bit}
\end{equation}
Note that if the representative bit is high (resp. low) then the augmented bit is also high (resp. low). The difference between the augmented and representative bit depends on the size of the available margin and the mean soft-bit value. Almost everywhere, an increase (resp. decrease) of the mean soft-bit increases (resp. decreases) the value of the augmented bit (see figure \ref{fig:margin-trick}). Note that if the $i$th bit is representative (i.e. hard-equivalent to the target function) then so is the augmented bit (see lemma \ref{prop:augmented}). We use margin packing, where appropriate, to define gradient-rich, hard-equivalents of boolean functions.

\subsection{Differentiable $\wedge$, $\vee$ and $\Rightarrow$}

We aim to construct a differentiable analogue of the boolean function $\bigwedge_{i=1}^{n} x_i$. A representative bit is $\operatorname{min}(x_{1},\dots,x_{n})$. The function
\begin{equation*}
\begin{aligned}
\partial_{\wedge}: [0,1]^{n} &\to [0,1], \\
{\bf x} &\mapsto \operatorname{augmented-bit}({\bf x}, \operatorname{argmin}\limits_{i} x[i])
\end{aligned}
\end{equation*}
is therefore hard-equivalent to the boolean function $\bigwedge_{i=1}^{n} x_i$ (see proposition \ref{prop:and}). In the special case $n=2$ we get the piecewise function,
\begin{equation*}
\partial_{\wedge}\!(x, y) =
	\begin{cases}
	1/2 + 1/2(x + y)(\operatorname{min}(x,y) - 1/2) & \text{if } \operatorname{min}(x,y) > 1/2 \\
	\operatorname{min}(x,y) + 1/2(x + y)(1/2 - \operatorname{min}(x,y)) & \text{otherwise.}
	\end{cases}
\end{equation*}
Note that $\partial_{\wedge}$ is differentiable almost everywhere and gradient-rich (see figure \ref{fig:gradient-rich}).

The differentiable analogue of $\vee$ is identical to $\wedge$, except the representative bit is selected by $\operatorname{max}$. The function
\begin{equation*}
\begin{aligned}
\partial_{\vee}: [0,1]^{n} &\to [0,1], \\
{\bf x} &\mapsto \operatorname{augmented-bit}({\bf x}, \operatorname{argmax}\limits_{i} x[i])
\end{aligned}
\end{equation*}
is hard-equivalent to the boolean function $\bigvee_{i=1}^{n} x_i$ (see proposition \ref{prop:or}). Note that $\partial_{\vee}$ is differentiable almost everywhere and gradient-rich (see figure \ref{fig:gradient-rich}).

The differentiable analogue of $\Rightarrow$ (material implication) is defined in terms of $\partial_{\vee}$. The function
\begin{equation*}
\begin{aligned}
\partial_{\Rightarrow}: [0,1]^{2} &\to [0,1],\\
(x, y) &\mapsto \partial_{\vee}\!(y, 1-x)\text{,}
\end{aligned}
\end{equation*}
is hard-equivalent to $x \Rightarrow y$ (see proposition \ref{prop:implies}). We can define  analogues of all the basic boolean operators in a similar manner.

\subsection{Differentiable majority}

\begin{figure}[t]
	\centering
	\includegraphics[trim=0pt 0pt 0pt 0pt, clip, width=1.0\textwidth]{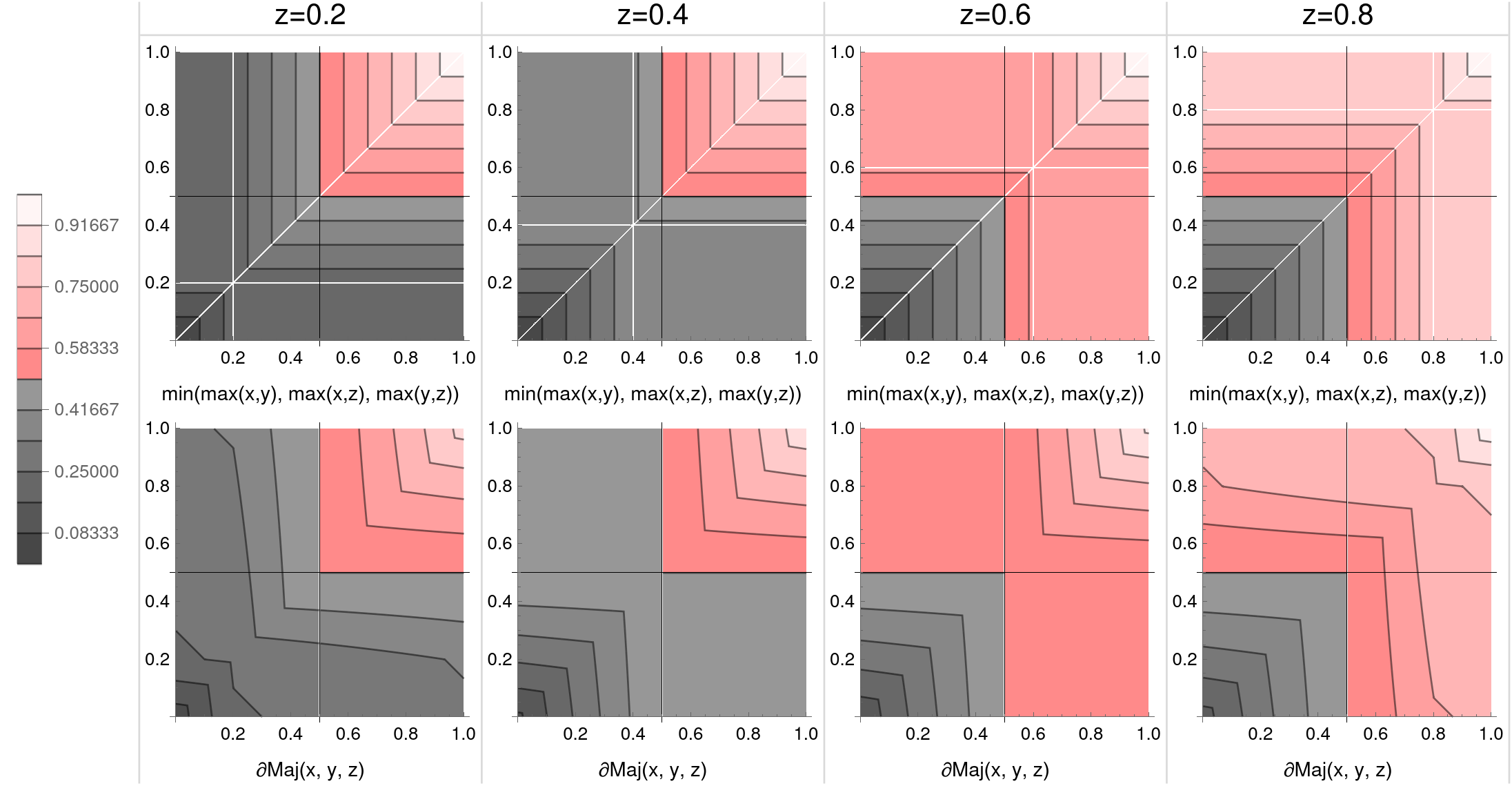}
	\caption{{\em Differentiable boolean majority.} The boolean majority function for three variables in DNF form is $\operatorname{Maj}(x,y,z) = (x \wedge y) \vee (x \wedge y) \vee (y \wedge z)$. The upper row contains contour plots of $f(x,y,z) = \operatorname{min}(\operatorname{max}(x,y), \operatorname{max}(x,z), \operatorname{max}(y,z))$ for values of $z \in \{0.2, 0.4, 0.6, 0.8\}$. $f$ is differentiable and $\btright\!\operatorname{Maj}$ but gradient-sparse (vertical and horizontal contours indicate constancy with respect to an input). Also, the number of terms in $f$ grows exponentially with the number of variables. The lower row contains contour plots of $\partial\!\operatorname{Maj}(x,y,z)$ for the same values of $z$. $\partial\!\operatorname{Maj}$ is differentiable and $\btright\!\operatorname{Maj}$ yet gradient-rich (curved contours indicate variability with respect to any inputs). In addition, the number of terms in $\partial\!\operatorname{Maj}$ is constant with respect to the number of variables.} 
	\label{fig:majority-plot}
\end{figure}

The boolean majority function is particularly important for tractable learning because it is a threshold function:
\begin{equation*}
\begin{aligned}
\operatorname{Maj}: \{0,1\}^{n} &\to \{0,1\},\\
{\bf x} &\mapsto \left\lfloor
\frac{1}{2} + \frac{\sum_{i=1}^{n} x_{i} - 1/2}{n}
\right\rfloor\text{,}
\end{aligned}
\end{equation*}
where we count $\operatorname{False}$ as $0$ and $\operatorname{True}$ as $1$. Interpret each input bit $x_{i}$ as a vote, yes or no, for a binary decision. If the majority of voters are in favour then $\operatorname{Maj}$ outputs 1. The majority function, in the context of a predictive model, aggregates multiple bits of weak evidence into a hard decision.We aim to construct a differentiable analogue of $\operatorname{Maj}$.

$\operatorname{Maj}$ for $n$ bits in DNF form is a disjunction of $\binom{n}{k}$ conjunctive clauses of size $k$, where $k=\lceil n/2 \rceil$. Each clause checks whether a unique combination of a majority of the $n$ bits are all high, e.g. $\operatorname{Maj}(x, y, z) = (x \wedge y) \vee (x \wedge y) \vee (y \wedge z)$. In principle we can implement a differentiable analogue of $\operatorname{Maj}$ in terms of $\partial_{\wedge}$ and $\partial_{\vee}$. However, the number of terms grows exponentially with the variables (e.g. $n=50$ generates over 100 trillion clauses, which is infeasible). And no general algorithm exists to find the minimal representation of $\operatorname{Maj}$ for arbitrary $n$.

Instead, we trade-off time for memory costs. Observe that if the function $\operatorname{sort}({\bf x})$ sorts the elements of ${\bf x}$ in ascending order then the `median' soft-bit is representative. For example, if ${\bf x} = [0.4, 0.9, 0.2]$ then $\operatorname{sort}({\bf x}) = [0.2, 0.4, 0.9]$ and the `median' bit $x_{2}=0.4$ is low, which is hard-equivalent to $\operatorname{Maj}(0, 1, 0) = 0$. Define the index of the `median' bit by
\begin{equation*}
\begin{aligned}
\operatorname{majority-index}: [0, 1]^{n} & \to \mathbb{Z}_{> 0}\\
{\bf x} & \mapsto \left\lceil \frac{|{\bf x}|}{2} \right\rceil
\text{.}
\end{aligned}
\end{equation*}
Then, applying margin packing, define the differentiable function
\begin{equation*}
\begin{aligned}
	\partial\!\operatorname{Maj}: [0,1]^{n} &\to [0,1], \\
	{\bf x} &\mapsto \operatorname{augmented-bit}(\operatorname{sort}({\bf x}), \operatorname{majority-index}({\bf x}))\text{,}
\end{aligned}
\end{equation*}
which is hard-equivalent to $\operatorname{Maj}$ (see theorem \ref{prop:majority}). Note that $\partial\!\operatorname{Maj}$ is differentiable almost everywhere and gradient-rich (see figure \ref{fig:majority-plot}). If $\operatorname{sort}$ is quicksort then the the average time-complexity of $\partial\!\operatorname{Maj}$ is $\mathcal{O}(n\log{}n)$, which makes $\partial\!\operatorname{Maj}$ more expensive than $\partial_{\neg}$, $\partial_{\wedge}$, $\partial_{\vee}$ and $\partial_{\Rightarrow}$ at training time. However, in the hard $\partial\mathbb{B}$ net we efficiently implement $\operatorname{Maj}$ as a discrete program that simply checks if the majority of bits are high. Note that we use sorting to define a differentiable function that is exactly equivalent to a discrete function (rather than defining a continuous approximation to sorting, e.g. \cite{NEURIPS2019_d8c24ca8}).

\subsection{Differentiable counting}

A boolean counting function $f({\bf x})$ is $\operatorname{True}$ if a counting predicate, $c({\bf x})$, holds over its $n$ inputs. We aim to construct a differentiable analogue of $\operatorname{count}({\bf x}, k)$ where $c({\bf x}) := |\{x_{i} : x_{i} = 1 \}| = k$ (i.e. `exactly $k$ high'), which can be useful in multiclass classification problems.

As before, we use $\operatorname{sort}$ to trade-off time for memory costs. Observe that if the elements of ${\bf x}$ are in ascending order then, if any soft-bits are high, there exists a unique contiguous pair of indices $(i,i+1)$ where $x_{i}$ is low and $x_{i+1}$ is high, where index $i$ is a direct count of the number of soft-bits that are low in ${\bf x}$. In consequence, define
\begin{equation*}
\begin{aligned}
\partial\!\operatorname{count-hot}: [0,1]^{n} &\to [0,1]^{n+1}, \\
{\bf x} &\mapsto \operatorname{low-high}(\operatorname{sort}({\bf x}))\text{,}
\end{aligned}
\end{equation*}
where 
\begin{equation*}
\begin{aligned}
\operatorname{low-high}: [0,1]^{n} &\to [0,1]^{n+1},\\
{\bf x} &\mapsto \left[ \partial_{\wedge}\!(1, x_{1}), \partial_{\wedge}\!(1 - x_{1}, x_{2}), \dots, \partial_{\wedge}\!(1 - x_{n-1}, x_{n}), \partial_{\wedge}\!(1-x_{n}, 1) \right]\text{.}
\end{aligned}
\end{equation*}
$\partial\!\operatorname{count-hot}({\bf x})$ outputs a 1-hot vector where the index of high bit is the number of low bits in ${\bf x}$. For example, $\partial\!\operatorname{count-hot}([0.1, 0.9, 0.2]) = [0.1, 0.2, \bold{0.8}, 0.1]\text{,}$ indicating that 2 bits are low, and $\partial\!\operatorname{count-hot}([0.6, 0.9, 0.7]) = [\bold{0.6}, 0.4, 0.3, 0.1]\text{,}$ indicating that 0 bits are low. Note that $\partial\!\operatorname{count-hot}$ is differentiable, gradient-rich and hard-equivalent to the boolean function
\begin{equation*}
\begin{aligned}
\operatorname{count-hot}: \{0,1\}^{n} &\to \{0,1\}^{n+1}, \\
{\bf x} &\mapsto \left[\operatorname{k-of-n}({\bf x}, 0), \operatorname{k-of-n}({\bf x}, 1), \dots, \operatorname{k-of-n}({\bf x}, n)\right]\text{,}
\end{aligned}
\end{equation*}
where
\begin{equation*}
\operatorname{k-of-n}({\bf x}, k) = \bigvee_{|S|=k} \bigwedge_{i\in S} x_i \bigwedge_{j\notin S} \neg x_j
\end{equation*}
(see proposition \ref{prop:count}). However, in the hard $\partial\mathbb{B}$ net we efficiently implement $\operatorname{count-hot}$ as a discrete program that simply counts the number of low bits.

We can construct various kinds of boolean counting functions from $\partial\!\operatorname{count-hot}$. For example, $\partial\!\operatorname{count}({\bf x}, k)$ is straightforwardly $\partial\!\operatorname{count-hot}({\bf x})[k]$ where we can use margin-packing to ensure that this single soft-bit is gradient-rich.

This basic set of boolean functions is sufficient to learn non-trivial relationships from data. We now turn to constructing $\partial\mathbb{B}$ nets from compositions of these functions.

\subsection{Boolean logic layers}

The fully variety of $\partial\mathbb{B}$ net architectures is to be explored. Here we focus on defining basic layers sufficient for the classification experiments in section \ref{sec:experiments}. Other kinds of layers, such as convolutional, or real encoders/decoders for regression problems, will be addressed in a sequel.

A $\partial_{\neg} \!\operatorname{Layer}$ of width $n$ learns to negate up to $n$ different subsets of the elements of its input vector:
\begin{equation*}
\begin{aligned}
\partial_{\neg} \!\operatorname{Layer}: [0,1]^{n \times m} \times [0,1]^{m} &\to [0,1]^{n \times m}, \\
({\bf W}, {\bf x}) &\mapsto 
\begin{bmatrix}
\partial_{\neg}(w_{1,1}, x_{1}) & \dots & \partial_{\neg}(w_{1,m}, x_{m}) \\
\vdots & \ddots & \vdots \\
\partial_{\neg}(w_{n,1}, x_{1}) & \dots & \partial_{\neg}(w_{n,m}, x_{m})
\end{bmatrix}
\end{aligned}
\end{equation*}
where ${\bf x}$ is a soft-bit input vector, ${\bf W}$ is a weight matrix and $n$ is the layer width. Similarly, A $\partial_{\Rightarrow} \!\operatorname{Layer}$ of width $n$ learns to `mask to true or $\operatorname{nop}$' up to $n$ different subsets of the elements of its input vector:
\begin{equation*}
\partial_{\Rightarrow} \!\operatorname{Layer}({\bf W}, {\bf x}) =
\begin{bmatrix}
\partial_{\Rightarrow}(w_{1,1}, x_{1}) & \dots & \partial_{\Rightarrow}(w_{1,m}, x_{m}) \\
\vdots & \ddots & \vdots \\
\partial_{\Rightarrow}(w_{n,1}, x_{1}) & \dots & \partial_{\Rightarrow}(w_{n,m}, x_{m})
\end{bmatrix}\text{.}
\end{equation*}
A $\partial_{\wedge}\!\operatorname{Neuron}$ learns to logically $\wedge$ a subset of its input vector:
\begin{equation*}
\begin{aligned}
\partial_{\wedge}\!\operatorname{Neuron}: [0,1]^{n} \times [0,1]^{n} &\to [0,1], \\
({\bf w}, {\bf x}) &\mapsto \min(\partial_{\Rightarrow}\!(w_{1}, x_{1}), \dots, \partial_{\Rightarrow}\!(w_{n}, x_{n}))\text{,}
\end{aligned}
\end{equation*}
where ${\bf w}$ is a weight vector. Each $\partial_{\Rightarrow}(w_{i},x_{i})$ learns to include or exclude $x_{i}$ from the conjunction depending on weight $w_{i}$. For example, if $w_{i}>0.5$ then $x_{i}$ affects the value of the conjunction since $\partial_{\Rightarrow}(w_{i},x_{i})$ passes-through a soft-bit that is high if $x_{i}$ is high, and low otherwise; but if $w_{i} \leq 0.5$ then $x_{i}$ does not affect the conjunction since $\partial_{\Rightarrow}(w_{i},x_{i})$ always passes-through a high soft-bit. A $\partial_{\wedge}\!\operatorname{Layer}$ of width $n$ learns up to $n$ different conjunctions of subsets of its input (of whatever size). A $\partial_{\vee}\!\operatorname{Neuron}$ is defined similarly:
\begin{equation*}
\begin{aligned}
\partial_{\vee}\!\operatorname{Neuron}: [0,1]^{n} \times [0,1]^{n} &\to [0,1], \\
({\bf w}, {\bf x}) &\mapsto \max(\partial_{\wedge}\!(w_{1}, x_{1}), \dots, \partial_{\wedge}\!(w_{n}, x_{n}))\text{.}
\end{aligned}
\end{equation*}
Each $\partial_{\wedge}(w_{i},x_{i})$ learns to include or exclude $x_{i}$ from the disjunction depending on weight $w_{i}$. A $\partial_{\vee}\!\operatorname{Layer}$ of width $n$ learns up to $n$ different disjunctions of subsets of its input (of whatever size).

We can compose $\partial_{\neg}$, $\partial_{\wedge}$ and $\partial_{\vee}$ layers to learn boolean formulae of arbitrary width and depth.

\subsection{Classification layers}

In classification problems the final layer of a neural network is typically interpreted as a vector of real-valued logits, one for each label, where the index of the maximum logit indicates the most probable label. However, we cannot interpret a soft-bit vector as logits without violating hard-equivalence. In addition, when training $\partial\mathbb{B}$ nets, loss functions should be a function of hardened bits, otherwise gradient descent may non-optimally traverse trajectories that take no account of the hard threshold at $1/2$. For example, consider that an instance is correctly classified by a 1-hot vector with high bit $x=0.51$. Updating the net's weights to change this value to $0.51+\epsilon$ will not improve accuracy and may prevent the correct classification of a different instance. 

For these reasons, $\partial\mathbb{B}$ nets have a final `hardening' layer to ensure that loss is a function of hard, not soft, bits:
\begin{equation*}
\begin{aligned}
\partial\!\operatorname{harden}: [0,1]^{n} &\to [0,1]^{n}, \\
{\bf x} &\mapsto \operatorname{harden}({\bf x})\text{.}
\end{aligned}
\end{equation*}
The $\operatorname{harden}$ function is not differentiable and therefore $\partial\!\operatorname{harden}$ uses the straight-through estimator \citep{DBLP:journals/corr/BengioLC13} during backpropagation. By restricting the use of the straight-through estimator to final layers we avoid compounding gradient estimation errors to deeper parts of the network. Note that $\partial\!\operatorname{harden}$ is hard-equivalent to a $\operatorname{nop}$.

$\partial\mathbb{B}$ nets can re-use many of the techniques deployed in standard neural networks. For example, for improved generalisation, we define a `boolean' analogue of the dropout layer \citep{JMLR:v15:srivastava14a}:
\begin{equation*}
\begin{aligned}
\partial\!\operatorname{dropout}: [0,1]^{n} \times [0,1] &\to [0,1]^{n}, \\
({\bf x}, p) &\mapsto [f(x_{1}, p), \dots, f(x_{n}, p)]\text{,}
\end{aligned}
\end{equation*}
where
\begin{equation*}
f(x, p) = \begin{cases}
1 - x, & \text{with probability } p \\
x, & \text{otherwise.}
\end{cases}
\end{equation*}
At train time $\partial\!\operatorname{dropout}$ randomly negates soft-bit values with probability $p$. At test time, and in the hard-net, $\partial\!\operatorname{dropout}$ is a $\operatorname{nop}$.

\begin{figure}[t!]
	\centering
	\includegraphics[width=0.6\textwidth]{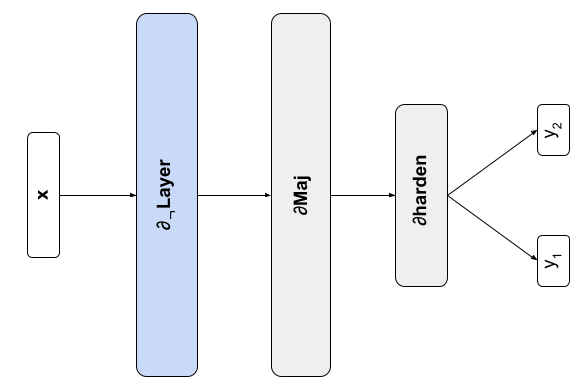}
	\caption{{\em A $\partial\mathbb{B}$ net to illustrate hardening}. The net concatenates a $\partial_{\neg}\!\operatorname{Layer}$ (of width $n$) with a reshaping layer that  outputs two vectors, which get reduced, by a $\partial\!\operatorname{Maj}$ operator,  to 2 soft-bits, one for each class label. A final $\partial\!\operatorname{harden}$ layer ensures the loss is a function of hard bits. When $|{\bf x}|=1$ we choose $n=2$ and therefore the net's weights, once hardened, consume $2$ bits. When $|{\bf x}|=5$ we choose $n=8$ and the weights consume $40$ bits ($5$ bytes).}
	\label{fig:toy-example-architecture}
\end{figure}

\section{Experiments}\label{sec:experiments}

The $\partial\mathbb{B}$ net library is implemented in Flax \citep{flax2020github} and JAX \citep{jax2018github} and available at {\small \url{github.com/Z80coder/db-nets}}. The library supports the specification of a $\partial\mathbb{B}$ net as Python code, which automatically defines (i) the soft-net for training (weights are floats), (ii) a hard-net for inference (weights are booleans), and (iii) a symbolic net for interpretation (weights and inputs are symbols). The symbolic net, when evaluated, interprets its own JAX expression and outputs a description of the discrete program it computes.

We compare the performance of $\partial\mathbb{B}$ nets against standard ML approaches on three problems: the classic Iris dataset, an adversarial noisy XOR problem, and MNIST. But first we illustrate the kind of discrete program that a $\partial\mathbb{B}$ net learns.

\subsection{Hardening}

We present a toy problem to illustrate hard-equivalence. Consider the trivial problem of predicting whether a person wears a $\operatorname{t-shirt}$ (label 0) or a $\operatorname{coat}$ (label 1) conditional on the single feature $\operatorname{outside}$ (0 = False, and 1 = True). The training and test data consist of the examples in table \ref{tab:toy1}.

\begin{table}[h!]
	\centering
	\begin{tabular}{|c|c|}
		$\operatorname{outside}$ & $\operatorname{label}$ \\ \hline
		0       & 0     \\
		1       & 1    
	\end{tabular}
	\caption{A trivial learning problem}
	\label{tab:toy1}
\end{table}

We use the $\partial\mathbb{B}$ net described in figure \ref{fig:toy-example-architecture}, which is hard-equivalent to the discrete program:

\begin{lstlisting}[language=Python,style=mystyle,frame=single]
def dbNet(outside):
  return [
    ge(sum((0, not(xor(ne(outside, 0), w1)))), 1),
	ge(sum((0, not(xor(ne(outside, 0), w2)))), 1)
  ]
\end{lstlisting}

with trainable weights $w_{1}$ and $w_{2}$. We randomly initialize the network and train using the RAdam optimizer \citep{Liu2020On} with softmax cross-entropy loss until training and test accuracies are both $100\%$. We harden the learned weights to get $w_{1} = \operatorname{False}$ and $w_{2} = \operatorname{True}$, and bind with the discrete program, which then symbolically simplifies to:

\begin{lstlisting}[language=Python,style=mystyle,frame=single]
def dbNet(outside):
	return [not(outside), outside]
\end{lstlisting}

which is directly interpretable as `when outside wear a $\operatorname{coat}$, otherwise wear a $\operatorname{t-shirt}$'.

Hardening scales to arbitrarily complex $\partial\mathbb{B}$ nets. Interpreting the net's predictions requires automatic symbolic simplification. For example, introduce 4 additional boolean features: $\operatorname{very-cold}$, $\operatorname{cold}$, $\operatorname{warm}$, and $\operatorname{very-warm}$. The training and test data consists of examples like those in table \ref{tab:toy2}.

\begin{table}[h!]
	\centering
	\begin{tabular}{|c|c|c|c|c|c|}
		$\operatorname{very-cold}$ & $\operatorname{cold}$ & $\operatorname{warm}$ & $\operatorname{very-warm}$ & $\operatorname{outside}$ & $\operatorname{label}$ \\ \hline
		1 & 0 & 0 & 0 & 0 & 1 \\
		0 & 0 & 0 & 1 & 1 & 1 \\
		0 & 0 & 1 & 0 & 1 & 0 \\
		0 & 0 & 0 & 1 & 0 & 0 \\
		\dots & \dots & \dots & \dots & \dots & \dots
	\end{tabular}
	\caption{A toy learning problem}
	\label{tab:toy2}
\end{table}

We use the same architecture but increase the width of the $\partial_{\neg}\!\operatorname{Layer}$ from 2 to 8. The net is now hard-equivalent to the discrete program:

\begin{lstlisting}[language=Python,style=mystyle,frame=single]
def dbNet(very-cold, cold, warm, very-warm, outside):
  return [
    ge(sum((sum((sum((sum((sum((sum((sum((sum((sum((sum((sum((sum((sum((sum((sum((sum((sum((sum((sum((sum((0, not(xor(ne(very-cold, 0), w1)))), not(xor(ne(cold, 0), w2)))), not(xor(ne(warm, 0), w3)))), not(xor(ne(very-warm, 0), w4)))), not(xor(ne(outside, 0), w5)))), not(xor(ne(very-cold, 0), w6)))), not(xor(ne(cold, 0), w7)))), not(xor(ne(warm, 0), w8)))), not(xor(ne(very-warm, 0), w9)))), not(xor(ne(outside, 0), w10)))), not(xor(ne(very-cold, 0), w11)))), not(xor(ne(cold, 0), w12)))), not(xor(ne(warm, 0), w13)))), not(xor(ne(very-warm, 0), w14)))), not(xor(ne(outside, 0), w15)))), not(xor(ne(very-cold, 0), w16)))), not(xor(ne(cold, 0), w17)))), not(xor(ne(warm, 0), w18)))), not(xor(ne(very-warm, 0), w19)))), not(xor(ne(outside, 0), w20)))), 11),
    ge(sum((sum((sum((sum((sum((sum((sum((sum((sum((sum((sum((sum((sum((sum((sum((sum((sum((sum((sum((sum((0, not(xor(ne(very-cold, 0), w21)))), not(xor(ne(cold, 0), w22)))), not(xor(ne(warm, 0), w23)))), not(xor(ne(very-warm, 0), w24)))), not(xor(ne(outside, 0), w25)))), not(xor(ne(very-cold, 0), w26)))), not(xor(ne(cold, 0), w27)))), not(xor(ne(warm, 0), w28)))), not(xor(ne(very-warm, 0), w29)))), not(xor(ne(outside, 0), w30)))), not(xor(ne(very-cold, 0), w31)))), not(xor(ne(cold, 0), w32)))), not(xor(ne(warm, 0), w33)))), not(xor(ne(very-warm, 0), w34)))), not(xor(ne(outside, 0), w35)))), not(xor(ne(very-cold, 0), w36)))), not(xor(ne(cold, 0), w37)))), not(xor(ne(warm, 0), w38)))), not(xor(ne(very-warm, 0), w39)))), not(xor(ne(outside, 0), w40)))), 11)
  ]
\end{lstlisting}
We train the $[w_{1}, \dots, w_{40}]$ soft-bit weights as before then harden to 40 boolean weights and bind with the discrete program. Post-training the program symbolically simplifies to:

\begin{lstlisting}[language=Python,style=mystyle,frame=single]
def dbNet(very-cold, cold, warm, very-warm, outside):
  return [
    4 !very-cold + 4 !cold + (3 warm + !warm) + (very-warm + 3 !very-warm) + (outside + 3 !outside) >= 11,
    (very-cold + 3 !very-cold) + 4 cold + 4 !warm + (3 very-warm + !very-warm) + 2 (outside + !outside) >= 11
  ]
\end{lstlisting}

The predictions linearly weight multiple pieces of evidence due to the presence of the $\partial\!\operatorname{Maj}$ operator (which is probably overkill for this toy problem). From this expression we can read-off that the $\partial\mathbb{B}$ net has learned `if not $\operatorname{very-cold}$ and not $\operatorname{cold}$ and not $\operatorname{outside}$ then wear a $\operatorname{t-shirt}$'; and `if $\operatorname{cold}$ and not ($\operatorname{warm}$ or $\operatorname{very-warm}$) and $\operatorname{outside}$ then wear a $\operatorname{coat}$' etc. The discrete program is more interpretable compared to typical neural networks, and can be exactly encoded as a SAT problem in order to verify its properties, such as robustness.

\begin{figure}[t!]
	\centering
	\includegraphics[width=0.8\textwidth]{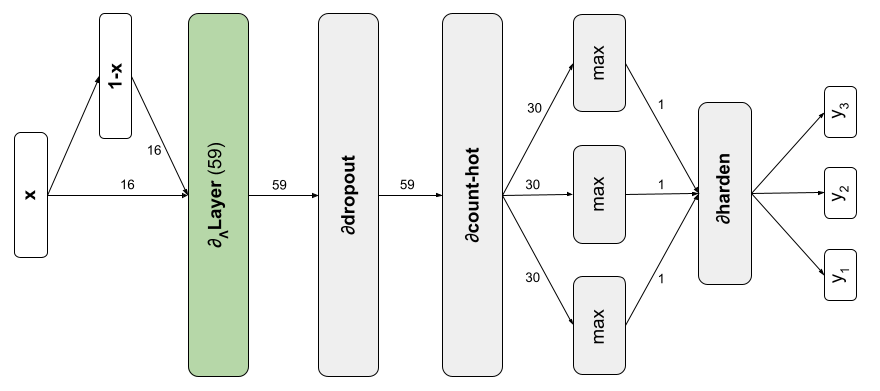}
	\caption{{\em A $\partial\mathbb{B}$ net for the binary Iris problem}. The net concatenates the soft-bit input, ${\bf x}$ (length 16), with its negation, ${\bf 1 - x}$, and supplies the resulting vector (length 32) to a $\partial_{\wedge}\!\operatorname{Layer}$ (width 59), a $\partial\!\operatorname{dropout}$ layer for improved generalisation, a $\partial\!\operatorname{count-hot}$ layer that generates a 1-hot vector (width 60) that is reduced by $\operatorname{max}$ to a 1-hot vector of 3 classification bits. A final $\partial\!\operatorname{harden}$ ensures the loss is a function of hard bits. The net's weights, once hardened, consume $236$ bytes.}
	\label{fig:binary-iris-architecture}
\end{figure}

\subsection{Binary Iris}

The Iris dataset has 150 examples with 4 inputs (sepal length and width, and petal length and width), and 3 labels ({\em setosa}, {\em versicolour}, and {\em virginica}). We use the binary version of the Iris dataset \citep{binary-iris-dataset} where each input float is represented by 4 bits. We perform 1000 experiments, each with a different random seed. Each experiment randomly partitions the data into 80\% training and 20\% test sets. We initialize the network, described in figure \ref{fig:binary-iris-architecture}, with all weights $w_{i} = 0.3$ and train for 1000 epochs with the RAdam optimizer and softmax cross-entropy loss. 

We measure the accuracy of the final net to avoid hand-picking the best configuration. Table \ref{tab:binary-iris-results} compares the $\delta\mathbb{B}$ net against other classifiers  \citep{granmo18}. Naive Bayes performs the worst. The Tsetlin machine performs best on this problem, with the $\partial\mathbb{B}$ net second.

\begin{table}[t]
	\centering
	\begin{tabular}{llllll}
		\cline{2-6}
		\multicolumn{1}{c}{}                       & \multicolumn{5}{c}{\textbf{accuracy}}                                                                                                                                                            \\ \cline{2-6} 
		\multicolumn{1}{l|}{}                      & \multicolumn{1}{l|}{mean}                  & \multicolumn{1}{l|}{5 \%ile}       & \multicolumn{1}{l|}{95 \%ile}       & \multicolumn{1}{l|}{min}           & \multicolumn{1}{l|}{max}            \\ \hline
		\multicolumn{1}{|l|}{Tsetlin}              & \multicolumn{1}{l|}{95.0 +/- 0.2}          & \multicolumn{1}{l|}{86.7}          & \multicolumn{1}{l|}{100.0}          & \multicolumn{1}{l|}{80.0}          & \multicolumn{1}{l|}{100.0}          \\ \hline
		\multicolumn{1}{|l|}{$\partial\mathbb{B}$} & \multicolumn{1}{l|}{\textbf{93.9 +/- 0.1}} & \multicolumn{1}{l|}{\textbf{86.7}} & \multicolumn{1}{l|}{\textbf{100.0}} & \multicolumn{1}{l|}{\textbf{80.0}} & \multicolumn{1}{l|}{\textbf{100.0}} \\ \hline
		\multicolumn{1}{|l|}{neural network}       & \multicolumn{1}{l|}{93.8 +/- 0.2}          & \multicolumn{1}{l|}{86.7}          & \multicolumn{1}{l|}{100.0}           & \multicolumn{1}{l|}{80.0}          & \multicolumn{1}{l|}{100.0}           \\ \hline
		\multicolumn{1}{|l|}{SVM}                  & \multicolumn{1}{l|}{93.6 +/- 0.3}          & \multicolumn{1}{l|}{86.7}          & \multicolumn{1}{l|}{100.0}           & \multicolumn{1}{l|}{76.7}          & \multicolumn{1}{l|}{100.0}           \\ \hline
		\multicolumn{1}{|l|}{naive Bayes}          & \multicolumn{1}{l|}{91.6 +/- 0.3}          & \multicolumn{1}{l|}{83.3}          & \multicolumn{1}{l|}{96.7}           & \multicolumn{1}{l|}{70.0}          & \multicolumn{1}{l|}{100.0}           \\ \hline
	\end{tabular}
	\caption{{\em Ranked binary Iris results}  measured over 1000 experiments.}
	\label{tab:binary-iris-results}
\end{table}

\begin{figure}[t!]
	\centering
	\includegraphics[width=0.8\textwidth]{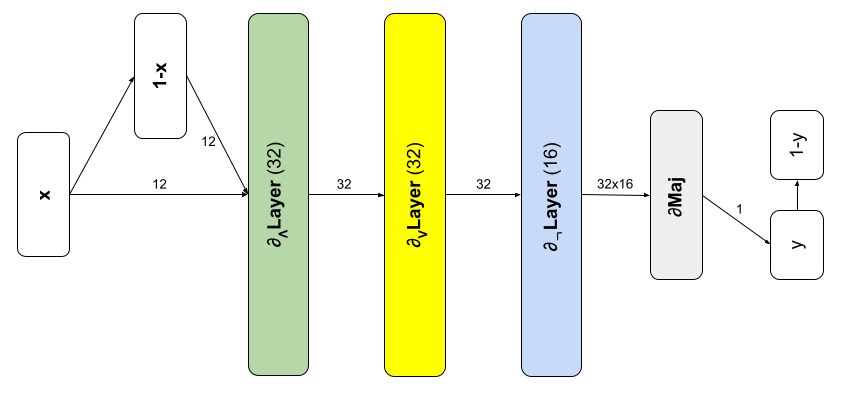}
	\caption{{\em A $\partial\mathbb{B}$ net for the noisy xor problem}. The net concatenates the soft-bit input, ${\bf x}$ (length 12), with its negation, ${\bf 1 - x}$, and supplies the resulting vector (length 24) to a $\partial_{\wedge}\!\!\operatorname{Layer}$ (width 32), $\partial_{\vee}\!\!\operatorname{Layer}$ (width 32),  $\partial_{\neg} \!\operatorname{Layer}$ (width 16), and a final $\partial\!\operatorname{Maj}$ to produce a single soft-bit $y \in [0,1]$ (to predict odd parity) and its negation $1-y$ (to predict even parity). The net's weights, once hardened, consume $288$ bytes.}
	\label{fig:noisy-xor-architecture}
\end{figure}

\subsection{Noisy XOR}

The noisy XOR dataset \citep{noisy-xor-dataset} is an adversarial parity problem with noisy non-informative features. The dataset consists of 10K examples with 12 boolean inputs and a target label (where 0 = odd and 1 = even) that is a XOR function of 2 of the inputs. The remaining 10 inputs are entirely random. We train on 50\% of the data where, additionally, 40\% of the labels are inverted. We initialize the network described in figure \ref{fig:noisy-xor-architecture} with random weights distributed close to the hard threshold at $1/2$ (i.e. in the $\partial_{\wedge}\!\operatorname{Layer}$, $w_{i} = 0.501 \times b + 0.3 \times (1-b)$ where $b \sim \operatorname{Bernoulli}(0.01)$; in the $\partial_{\vee}\!\operatorname{Layer}$, $w_{i} = 0.7 \times b + 0.499 \times (1-b)$ where $b \sim \operatorname{Bernoulli}(0.99)$); and in the $\partial_{\neg}\!\operatorname{Layer}$, $w_{i} \sim \operatorname{Uniform}(0.499, 0.501)$. We train for 2000 epochs with the RAdam optimizer and softmax cross-entropy loss. 

We measure the accuracy of the final net on the test data to avoid hand-picking the best configuration. Table \ref{tab:noisy-xor-results} compares the $\partial\mathbb{B}$ net against other classifiers \citep{granmo18}. The high noise causes logistic regression and naive Bayes to randomly guess. The SVM hardly performs better. In constrast, the multilayer neural network, Tsetlin machine, and  $\partial\mathbb{B}$ net all successfully learn the underlying XOR signal. The Tsetlin machine performs best on this problem, with the $\partial\mathbb{B}$ net second.

\begin{table}[t]
	\centering
	\begin{tabular}{llllll}
		\cline{2-6}
		\multicolumn{1}{c}{}                       & \multicolumn{5}{c}{\textbf{accuracy}}                                                                                                                                                            \\ \cline{2-6} 
		\multicolumn{1}{l|}{}                      & \multicolumn{1}{l|}{mean}                  & \multicolumn{1}{l|}{5 \%ile}       & \multicolumn{1}{l|}{95 \%ile}       & \multicolumn{1}{l|}{min}           & \multicolumn{1}{l|}{max}            \\ \hline
		\multicolumn{1}{|l|}{Tsetlin}              & \multicolumn{1}{l|}{99.3 +/- 0.3}          & \multicolumn{1}{l|}{95.9}          & \multicolumn{1}{l|}{100.0}          & \multicolumn{1}{l|}{91.6}          & \multicolumn{1}{l|}{100.0}          \\ \hline
		\multicolumn{1}{|l|}{$\partial\mathbb{B}$} & \multicolumn{1}{l|}{\textbf{97.9 +/- 0.2}} & \multicolumn{1}{l|}{\textbf{95.4}} & \multicolumn{1}{l|}{\textbf{100.0}} & \multicolumn{1}{l|}{\textbf{93.6}} & \multicolumn{1}{l|}{\textbf{100.0}} \\ \hline
		\multicolumn{1}{|l|}{neural network}       & \multicolumn{1}{l|}{95.4 +/- 0.5}          & \multicolumn{1}{l|}{90.1}          & \multicolumn{1}{l|}{98.6}           & \multicolumn{1}{l|}{88.2}          & \multicolumn{1}{l|}{99.9}           \\ \hline
		\multicolumn{1}{|l|}{SVM}                  & \multicolumn{1}{l|}{58.0 +/- 0.3}          & \multicolumn{1}{l|}{56.4}          & \multicolumn{1}{l|}{59.2}           & \multicolumn{1}{l|}{55.4}          & \multicolumn{1}{l|}{66.5}           \\ \hline
		\multicolumn{1}{|l|}{naive Bayes}          & \multicolumn{1}{l|}{49.8 +/- 0.2}          & \multicolumn{1}{l|}{48.3}          & \multicolumn{1}{l|}{51.0}           & \multicolumn{1}{l|}{41.3}          & \multicolumn{1}{l|}{52.7}           \\ \hline
		\multicolumn{1}{|l|}{logistic regression}  & \multicolumn{1}{l|}{49.8 +/- 0.3}          & \multicolumn{1}{l|}{47.8}          & \multicolumn{1}{l|}{51.1}           & \multicolumn{1}{l|}{41.1}          & \multicolumn{1}{l|}{53.1}           \\ \hline
	\end{tabular}
	\caption{{\em Ranked noisy XOR results}  measured over 100 experiments.}
	\label{tab:noisy-xor-results}
\end{table}

\begin{figure}[t!]
	\centering
	\includegraphics[width=0.7\textwidth]{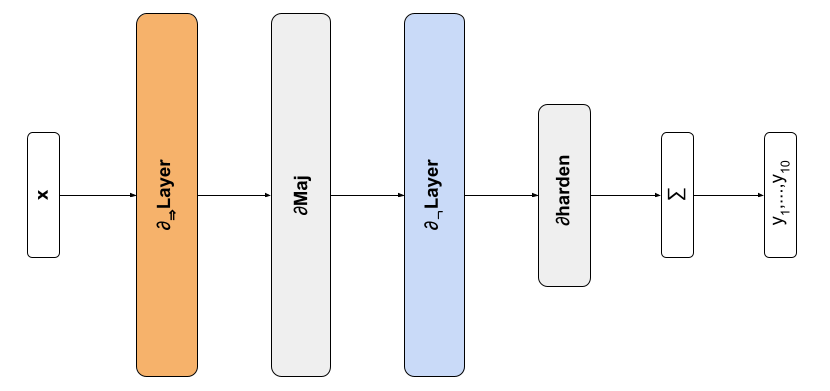}
	\caption{{\em A non-convolutional $\partial\mathbb{B}$ net for MNIST}. The input is a $28\times28$ bit matrix representing an image. The net consists of a $\partial_{\Rightarrow}\!\operatorname{Layer}$ (of width 60, to produce a $2940\times16$ reshaped array), a $\partial\!\operatorname{Maj}$ layer (to produce a vector of size $2940$), a $\partial_{\neg}\!\operatorname{Layer}$ (of width 20, to produce a $20 \times 2940$ array), and a final $\partial\!\operatorname{harden}$ operator to generate hard-bits split into 10 buckets and summed to produce 10 integer logits. The net's weights, once hardened, consume 13.23 kb.}
	\label{fig:mnist-architecture}
\end{figure}

\subsection{MNIST}

\begin{table}[t]
	\centering
	\begin{tabular}{lc}
		\cline{2-2}
		& \textbf{accuracy}                  \\ \hline
		\multicolumn{1}{|l|}{\em 2-layer NN, 800 HU, cross-entropy loss} & \multicolumn{1}{c|}{98.6} \\ \hline
		\multicolumn{1}{|l|}{Tsetlin}                        & \multicolumn{1}{c|}{98.2 +/- 0.0}  \\ \hline
		\multicolumn{1}{|l|}{\em K-nearest-neighbours, L3}       & \multicolumn{1}{c|}{97.2}          \\ \hline
		\multicolumn{1}{|l|}{$\partial\mathbb{B}$}           & \multicolumn{1}{c|}{\textbf{94.0}} \\ \hline
		\multicolumn{1}{|l|}{Logistic regression}            & \multicolumn{1}{c|}{91.5}          \\ \hline
		\multicolumn{1}{|l|}{\em Linear classifier (1-layer NN)} & \multicolumn{1}{c|}{88.0}          \\ \hline
		\multicolumn{1}{|l|}{Decision tree}                  & \multicolumn{1}{c|}{87.8}          \\ \hline
		\multicolumn{1}{|l|}{Multinomial Naive Bayes}        & \multicolumn{1}{c|}{83.2}          \\ \hline
	\end{tabular}
	\caption{{\em Ranked MNIST results}. A classifier in {\em italics} was trained on grey-value pixel data, otherwise the classifier was trained on binarized data. Note: the $\partial\mathbb{B}$ results are from a small model that under-fits the data (due to OOM errors on my GPU). The next draft will include results using a larger $\partial\mathbb{B}$ net.}
	\label{tab:mnist-table}
\end{table}

The MNIST dataset \citep{726791} consists of 60K training and 10K test examples of handwritten digits (0-9). We binarize the data by replacing pixels with grey value greater than 0.3 with 1, otherwise with 0. We initialize the network described in figure \ref{fig:mnist-architecture} with random weights distributed as $w_{i} = 0.501 \times b + 0.3 \times (1-b)$ where $b \sim \operatorname{Bernoulli}(0.01)$. We train for 1000 epochs with a batch size of 6000 using the RAdam optimizer and softmax cross-entropy loss.

We measure the accuracy on the final net. Table \ref{tab:mnist-table} compares the $\partial\mathbb{B}$ net against other classifiers (reference data taken from \cite{granmo18} and \url{yann.lecun.com/exdb/mnist}). Basic versions of the algorithms (e.g. no convolutional nets) are applied to unenhanced data (e.g. no data augmentation). The aim is to compare raw performance rather than optimise for MNIST. A 2-layer neural network trained on grey-value pixel data performs best. A Tsetlin machine of 40,000 automata each with 256 states (and therefore 40 kb of parameters) trained on binary data achieves $\approx 98.2\%$ accuracy. A $\partial\mathbb{B}$ net with 105,840 soft-bit weights that harden to 1-bit booleans (and therefore 13.23 kb of parameters) trained on binary data achieves $\approx 94.0\%$ accuracy. However, this $\partial\mathbb{B}$ net underfits the training data and we expect better performance from a larger model.

\section{Conclusion}\label{sec:conclusion}

$\partial\mathbb{B}$ nets are differentiable neural networks that are hard-equivalent to non-differentiable, boolean-valued functions. $\partial\mathbb{B}$ nets can therefore learn discrete functions by gradient descent. The main novelty of $\partial\mathbb{B}$ nets is the semantic equivalence between their two aspects: a differentiable soft-net and a non-differentiable hard-net. Maintaining this semantic equivalence requires defining new kinds of differentiable functions that are hard-equivalent to boolean functions, such as non-differentiable boolean majority. We propose `margin packing' as a potentially general technique for constructing differentiable functions that are hard-equivalent yet gradient-rich (and therefore backpropagate error to all their inputs). An advantage of $\partial\mathbb{B}$ nets is that we train the soft-net using efficient backpropagation on GPUs then `harden' to generate a learned discrete function that, unlike existing approaches to neural network binarization, has provably identical accuracy.

$\partial\mathbb{B}$ nets, being ultimately of a discrete and logical nature, are easier to interpret compared to standard neural networks, for example generating propositional formulae that can be further analysed, either by symbolic simplification or verification by SAT solvers. These properties are important in safety-critical domains. In addition, $\partial\mathbb{B}$ nets at inference time are highly compact, due to 1-bit weights, and potentially cheap to evaluate, as they reduce to bit manipulation and integer arithmetic. These properties are important in resource-poor deployment environments, such as edge devices. Further, due to the differentiable nature of $\partial\mathbb{B}$ nets, they can be arbitrarily composed with standard neural nets (e.g. by embedding them within standard nets to introduce domain-specific logical bias).

Preliminary experiments on three classification benchmarks demonstrate that $\partial\mathbb{B}$ nets can outperform multilayer perceptron networks, support vector machines, decision trees, and logistic regression. In terms of classification accuracy, the non-differentiable Tsetlin machine outperforms $\partial\mathbb{B}$ nets, which indicates room for futher improvements, e.g. by defining more expressive $\partial\mathbb{B}$ net layers (threshold functions with a learnable integer threshold, boolean decision lists etc.) and architectures (convolutional, regression nets, skip connections, attention etc.). In other words, this paper is only a first step towards exploring the space of differentiable nets that satisfy the requirement of hard-equivalence.

\subsubsection*{Acknowledgments}
Thanks to GitHub Next for sponsoring this research. And thanks to Pavel Augustinov, Richard Evans, Johan Rosenkilde, Max Schaefer, Ganesh Sittampalam, Tam\'{a}s Szab\'{o} and Albert Ziegler for helpful discussions and feedback.

\bibliographystyle{iclr2021_conference}
\bibliography{db}

\appendix

\section*{Appendix}

\section{Proofs}

\begin{prop}\label{prop:not}
	$\partial_{\neg}(x,y) \btright \neg (x \oplus y)$.
\begin{proof}
	Table \ref{not-table} is the truth table of the boolean function $\neg (x \oplus w)$, where $h(x) = \operatorname{harden}(x)$.
	\begin{table}[h!]
		\begin{center}
			\begin{tabular}{ccccccc}
				\multicolumn{1}{c}{$x$}  &\multicolumn{1}{c}{$y$}  &\multicolumn{1}{c}{$h(x)$}  &\multicolumn{1}{c}{$h(y)$} &\multicolumn{1}{c}{$\partial_{\neg}(x, y)$} &\multicolumn{1}{c}{$h(\partial_{\neg}(x, y))$}
				&\multicolumn{1}{c}{$\neg (h(y) \oplus h(x))$}
				\\ \hline \\
				$\left[0, \frac{1}{2}\right)$ & $\left[0, \frac{1}{2}\right)$ & 0 & 0 & $\left(\frac{1}{2},1\right]$ & 1 & 1\\[0.1cm] 
				$\left(\frac{1}{2}, 1\right]$ & $\left[0, \frac{1}{2}\right)$ &1 & 0 & $\left[0, \frac{1}{2}\right)$ & 0 & 0\\[0.1cm]
				$\left[0, \frac{1}{2}\right)$ & $\left(\frac{1}{2}, 1\right]$ &0 & 1 & $\left[0, \frac{1}{2}\right)$ & 0 & 0\\[0.1cm]
				$\left(\frac{1}{2}, 1\right]$ & $\left(\frac{1}{2}, 1\right]$ &1 & 1 & $\left(\frac{1}{2}, 1\right]$ & 1 & 1\\[0.1cm]
			\end{tabular}
		\end{center}
		\caption{$\partial_{\neg}(x,y) \btright \neg (y \oplus x)$.}\label{not-table}
	\end{table}
\end{proof}
\end{prop}

\begin{lemma}\label{prop:augmented}
	If a representative bit, $x_{i}$, is hard-equivalent to a target function, $g$, then so is the augmented bit, $z$.
	\begin{proof}
		As $x_{i}$ is representative then $\operatorname{harden}(x_{i}) = g(\operatorname{harden}({\bf x}))$. The augmented bit, $z$, is given by  \eqref{eq:augmented-bit}:
		\begin{equation*}
		z = \begin{cases}
		1/2 + \bar{\bf x}\times|x_{i} - 1/2| & \text{if } x_{i} > 1/2 \\
		x_{i} + \bar{\bf x}\times|x_{i} - 1/2| & \text{otherwise.}
		\end{cases}
		\end{equation*}
		In consequence,
		\begin{equation*}
		\operatorname{harden}(z) = \begin{cases}
		1 & \text{if } x > 1/2 \\
		0 & \text{otherwise,}
		\end{cases}
		\end{equation*}
		since $x_{i} > 1/2 \Rightarrow z > 1/2$ and $x_{i} \leq 1/2 \Rightarrow z \leq 1/2$. Hence, $\operatorname{harden}(z) = \operatorname{harden}(x_{i}) = g(\operatorname{harden}({\bf x}))$
	\end{proof}
\end{lemma}

\begin{prop}\label{prop:and}
	$\partial_{\wedge}\!(x,y) \btright x \wedge y$.
\begin{proof}
	Table \ref{and-table} is the truth table of the boolean function $x \wedge y$, where $h(x) = \operatorname{harden}(x)$..
	\begin{table}[h!]
		\begin{center}
			\begin{tabular}{ccccccc}
				\multicolumn{1}{c}{$x$}  &\multicolumn{1}{c}{$y$}  &\multicolumn{1}{c}{$h(x)$}  &\multicolumn{1}{c}{$h(y)$} &\multicolumn{1}{c}{$\partial_{\wedge}(x, y)$} &\multicolumn{1}{c}{$h(\partial_{\wedge}(x, y))$}
				&\multicolumn{1}{c}{$h(x) \wedge h(y)$}
				\\ \hline \\
				$\left[0, \frac{1}{2}\right)$ & $\left[0, \frac{1}{2}\right)$ & 0 & 0 & $\left[0, \frac{1}{2}\right)$ & 0 & 0\\[0.1cm]
				$\left(\frac{1}{2}, 1\right]$ & $\left[0, \frac{1}{2}\right)$ &1 & 0 & $\left(\frac{1}{4}, \frac{1}{2}\right)$ & 0 & 0\\[0.1cm]
				$\left[0, \frac{1}{2}\right)$ & $\left(\frac{1}{2}, 1\right]$ &0 & 1 & $\left(\frac{1}{4}, \frac{1}{2}\right)$ & 0 & 0\\[0.1cm]
				$\left(\frac{1}{2}, 1\right]$ & $\left(\frac{1}{2}, 1\right]$ &1 & 1 & $\left(\frac{1}{2}, 1\right]$ & 1 & 1\\[0.1cm]
			\end{tabular}
		\end{center}
		\caption{$\partial_{\wedge}(x,y) \btright x \wedge y$.}\label{and-table}
	\end{table}			
\end{proof}
\end{prop}

\begin{prop}\label{prop:or}
	$\partial_{\vee}\!(x,y) \btright x \vee y$.
\begin{proof}
	Table \ref{or-table} is the truth table of the boolean function $x \vee y$, where $h(x) = \operatorname{harden}(x)$..
	\begin{table}[h!]
	\begin{center}
		\begin{tabular}{ccccccc}
			\multicolumn{1}{c}{$x$}  &\multicolumn{1}{c}{$y$}  &\multicolumn{1}{c}{$h(x)$}  &\multicolumn{1}{c}{$h(y)$} &\multicolumn{1}{c}{$\partial_{\vee}(x, y)$} &\multicolumn{1}{c}{$h(\partial_{\vee}(x, y))$}
			&\multicolumn{1}{c}{$h(x) \vee h(y)$}
			\\ \hline \\
			$\left[0, \frac{1}{2}\right)$ & $\left[0, \frac{1}{2}\right)$ & 0 & 0 & $\left[0,\frac{1}{2}\right)$ & 0 & 0\\[0.1cm]
			$\left(\frac{1}{2}, 1\right]$ & $\left[0, \frac{1}{2}\right)$ &1 & 0 & $\left(\frac{1}{2},1\right]$ & 1 & 1\\[0.1cm]
			$\left[0, \frac{1}{2}\right)$ & $\left(\frac{1}{2}, 1\right]$ &0 & 1 & $\left(\frac{1}{2},1\right]$ & 1 & 1\\[0.1cm]
			$\left(\frac{1}{2}, 1\right]$ & $\left(\frac{1}{2}, 1\right]$ &1 & 1 & $\left(\frac{1}{2},1\right]$ & 1 & 1\\[0.1cm]
		\end{tabular}
	\end{center}
	\caption{$\partial_{\vee}(x,y) \btright x \vee y$.}\label{or-table}
	\end{table}			
\end{proof}
\end{prop}

\begin{prop}\label{prop:implies}
	$\partial_{\Rightarrow}\!(x,y) \btright x \Rightarrow y$.
\begin{proof}
	Table \ref{implies-table} is the truth table of the boolean function $x \Rightarrow y$, where $h(x) = \operatorname{harden}(x)$..
	\begin{table}[h!]
	\begin{center}
		\begin{tabular}{ccccccc}
			\multicolumn{1}{c}{$x$}  &\multicolumn{1}{c}{$y$}  &\multicolumn{1}{c}{$h(x)$}  &\multicolumn{1}{c}{$h(y)$} &\multicolumn{1}{c}{$\partial_{\Rightarrow}(x, y)$} &\multicolumn{1}{c}{$h(\partial_{\Rightarrow}(x, y))$}
			&\multicolumn{1}{c}{$h(x) \Rightarrow h(y)$}
			\\ \hline \\
			$\left[0, \frac{1}{2}\right)$ & $\left[0, \frac{1}{2}\right)$ & 0 & 0 & $\left(\frac{1}{2}, 1\right]$ & 1 & 0\\[0.1cm]
			$\left(\frac{1}{2}, 1\right]$ & $\left[0, \frac{1}{2}\right)$ &1 & 0 & $\left[0, \frac{1}{2}\right)$ & 0 & 0\\[0.1cm]
			$\left[0, \frac{1}{2}\right)$ & $\left(\frac{1}{2}, 1\right]$ &0 & 1 & $\left(\frac{1}{2},1\right]$ & 1 & 0\\[0.1cm]
			$\left(\frac{1}{2}, 1\right]$ & $\left(\frac{1}{2}, 1\right]$ &1 & 1 & $\left(\frac{1}{2}, \frac{7}{8}\right)$ & 1 & 0\\[0.1cm]
		\end{tabular}
	\end{center}
	\caption{$\partial_{\Rightarrow}(x,y) \btright x \Rightarrow y$.}\label{implies-table}
	\end{table}			
\end{proof}
\end{prop}

\begin{lemma}
\label{lem:maj}
Let $i$ = $\operatorname{majority-index}({\bf x})$, then the $i$th element of $\operatorname{sort}({\bf x})$ is hard-equivalent to boolean majority, i.e. $\operatorname{harden}(\operatorname{sort}({\bf x})[i]) = \operatorname{Maj}(\operatorname{harden}({\bf x}))$.
\begin{proof}
Let $h$ denote the number of bits that are high in ${\bf x} = [x_{1}, \dots, x_{n}]$. Then indices $\{j : n-h+1 \leq j \leq n\}$ are high in $\operatorname{sort}({\bf x})$. If the majority of bits are high, $h \geq \lfloor n/2 + 1 \rfloor$, then index $j=n - \lfloor n/2 + 1 \rfloor + 1 = n - \lfloor n/2 \rfloor = \lceil n/2 \rceil$ is high in $\operatorname{sort}({\bf x})$. $\operatorname{majority-index}$ selects index $i = \lceil n/2 \rceil$ and therefore $i=j$. Hence, if the majority of bits are high then $\operatorname{sort}({\bf x})[i]$ is high. Similarly, if the majority of bits are low, $h < \lfloor n/2 + 1 \rfloor$, then index $j=n - \lfloor n/2 + 1 \rfloor + 1 = n - \lfloor n/2 \rfloor = \lceil n/2 \rceil$ is low in $\operatorname{sort}({\bf x})$. Hence, if the majority of bits are low then $\operatorname{sort}({\bf x})[i]$ is low.

Note that $h \geq \lfloor n/2 + 1 \rfloor$ implies that $\operatorname{Maj}(\operatorname{harden}({\bf x})) \geq \left\lfloor \frac{1}{2} + \frac{1}{n}\left(\frac{n}{2} + 1 - \frac{1}{2} \right) \right\rfloor \geq \left\lfloor 1 + \frac{1}{2n} \right\rfloor = 1$, and $h < \lfloor n/2 + 1 \rfloor$ implies that $\operatorname{Maj}(\operatorname{harden}({\bf x})) < \left\lfloor 1 + \frac{1}{2n} \right\rfloor = 0$.

In consequence, $\operatorname{harden}(\operatorname{sort}({\bf x})[i]) = \operatorname{Maj}(\operatorname{harden}({\bf x}))$ for all $h \in [0,\dots, n]$.
\end{proof}
\end{lemma}

\begin{theorem}\label{prop:majority}
	$\partial\!\operatorname{Maj} \btright \operatorname{Maj}$.
\begin{proof}
	$\partial\!\operatorname{Maj}$ augments the representative bit $x_{i} = \operatorname{sort}({\bf x})[\operatorname{majority-index}({\bf x})]$. By lemma \ref{lem:maj} the representative bit is $\btright \operatorname{Maj}(\operatorname{harden}({\bf x}))$.
    By lemma \ref{prop:augmented}, the augmented bit, $\operatorname{augmented-bit}(\operatorname{sort}({\bf x}), \operatorname{majority-index}({\bf x}))$, is also $\btright\!\operatorname{Maj}(\operatorname{harden}({\bf x}))$. Hence $\partial\!\operatorname{Maj} \btright\!\operatorname{Maj}$.
\end{proof}
\end{theorem}

\begin{prop}\label{prop:count}
	$\partial\!\operatorname{count-hot} \btright \operatorname{count-hot}$.
	\begin{proof}
	Let $l$ denote the number of bits that are low in ${\bf x} = [x_{1},\dots,x_{n}]$, and let ${\bf y} = \partial\!\operatorname{count-hot}({\bf x})$. Then ${\bf y}[l+1]$ is high and any ${\bf y}[i]$, where $i \neq l+1$, is low. Let ${\bf z} = \operatorname{count-hot}(\operatorname{harden}({\bf x}))$. Then ${\bf z}[l+1]$ is high and any ${\bf z}[i]$, where $i \neq l+1$, is low. Hence, $\operatorname{harden}({\bf y}) = {\bf z}$, and therefore $\partial\!\operatorname{count-hot} \btright \operatorname{count-hot}$.
	\end{proof}
\end{prop}

\end{document}